\documentclass{article}

\PassOptionsToPackage{numbers}{natbib}
\usepackage[preprint]{neurips_2023}

\usepackage[utf8]{inputenc} % allow utf-8 input
\usepackage[T1]{fontenc}    % use 8-bit T1 fonts
\usepackage{hyperref}       % hyperlinks
\usepackage{url}            % simple URL typesetting
\usepackage{booktabs}       % professional-quality tables
\usepackage{amsfonts}       % blackboard math symbols
\usepackage{nicefrac}       % compact symbols for 1/2, etc.
\usepackage{microtype}      % microtypography
\usepackage{xcolor}         % colors
\usepackage{epsfig}
\usepackage{graphicx}
\usepackage{amsmath}
\usepackage{amssymb}
\usepackage{subcaption}
\usepackage{algorithm}
\usepackage{algpseudocode}
\usepackage{pifont}

\title{FedSDD: Scalable and Diversity-enhanced Distillation for Model Aggregation in Federated Learning}

\author{%
  Ho Man Kwan, Shenghui Song \\
  \texttt{hmkwan@connect.ust.hk}, \texttt{eeshsong@ust.hk} \\
  The Hong Kong University of Science and Technology \\
}

\begin{document}

\maketitle

\begin{abstract}
    Recently, innovative model aggregation methods based on knowledge distillation (KD) have been proposed for federated learning (FL). These methods not only improved the robustness of model aggregation over heterogeneous learning environment, but also allowed training heterogeneous models on client devices. However, the scalability of existing methods is not satisfactory, because the training cost on the server increases with the number of clients, which limits their application in large scale systems. Furthermore, the ensemble of existing methods is built from a set of client models initialized from the same checkpoint, causing low diversity. In this paper, we propose a scalable and diversity-enhanced federated distillation scheme, FedSDD, which decouples the training complexity from the number of clients to enhance the scalability, and builds the ensemble from a set of aggregated models with enhanced diversity. In particular, the teacher model in FedSDD is an ensemble built by a small group of aggregated (global) models, instead of all client models, such that the computation cost will not scale with the number of clients. Furthermore, to enhance diversity, FedSDD only performs KD to enhance one of the global models, i.e., the \textit{main global model}, which improves the performance of both the ensemble and the main global model. While partitioning client model into more groups allow building an ensemble with more aggregated models, the convergence of individual aggregated models will be slow down. We introduce the temporal ensembling which leverage the issues, and provide significant improvement with the heterogeneous settings. Experiment results show that FedSDD outperforms other FL methods, including FedAvg and FedDF, on the benchmark datasets.
\end{abstract}

%-------------------------------------------------------------------------

\section{Introduction}
Federated learning (FL) \citep{mcmahan2017communication} allows users to jointly train a deep learning model without sharing their data. Recent works \citep{lin2020ensemble, zhang2022fine, huang2022learn, cho2022heterogeneous} adopted knowledge distillation (KD) \citep{hinton2015distilling} for model aggregation to tackle data and device heterogeneity issues. Such distillation-based aggregation methods have been shown to outperform weight averaging methods like FedAvg \citep{mcmahan2017communication}. However, existing works utilized all client models to build the ensemble for KD, leading to poor scalability in real world applications, where a large number, e.g., thousands, of clients are available. Furthermore, existing methods built the ensemble by client models that were initialized from the same global model, which may have low diversity and thus limit the performance of the ensemble.

To tackle the above issues, we propose a new algorithm named Federated Scalable and Diversity-enhanced Distillation (FedSDD). The key innovations of FedSDD include: 
(i). Improving the scalability of distillation-based aggregation by building the ensemble (i.e. the teacher model) from a set of aggregated models, each of which is obtained by performing weight averaging over a sub-group of client models. As a result, the computation complexity for distillation at the server is independent of the number of clients but dictated by the number of aggregated models.
(ii). Maximizing the capacity of the ensemble and the global model by enhancing the diversity among the compositing models of the ensemble. This is achieved by training multiple global models simultaneously, but only transferring the knowledge from the ensemble to one of them. 
(iii). Ideally, a larger ensemble, i.e., more involved models, has better capacity. However, in the scheme of leveraging aggregated models as an ensemble ((i).), this results in fewer client models per group, implicating a trade-off between individual global model convergence and ensemble capacity. We propose alleviating this issue by introducing the temporal ensembling.
Furthermore, FedSDD has favorable properties in terms of efficiency and privacy. In particular, FedSDD can achieve higher efficiency by parallel training across the server and client sides. Different from existing distillation-based methods, FedSDD does not require accessing client models, which provides further protection to the user privacy. Experiment results show that FedSDD can achieve the superior results on CIFAR10/100 datasets \citep{krizhevsky2009learning}, especially when the data of different clients are not independent and identically distributed (Non-IID).

The contributions of this paper include:

1.) We propose FedSDD, a highly efficient and scalable distillation-based FL algorithm, to tackle the limitation of existing FL algorithms from the perspective of model aggregation. Building the ensemble by aggregated models over a group of clients improves the scalability while distillation to only one of the group models enhances the diversity among the compositing models of the ensemble. 

2.) Experiment results demonstrate that FedSDD can achieve superior performance with lower complexity and latency compared to existing distillation-based FL algorithms. Ablation studies highlight the impact of the key elements in FedSDD, which can potentially benefit further works along the same line of research.

%-------------------------------------------------------------------------

\section{Related works}

\textbf{Federated learning.} Training deep learning models require a large amount of data. In many practical scenarios, e.g. healthcare, data are distributed over the user devices, and these user data may not be able to be shared due to privacy regulations and/or communication constraints. To tackle the above issues, federated learning \citep{mcmahan2017communication} was proposed to train a shared model, normally referred to as the global model, by utilizing the data distributed on the client devices without data sharing. The simplest approach is FedAvg \citep{mcmahan2017communication}, which performs multiple epochs of local training on the client side and then aggregates the learned models on the server side by weight averaging. FedProx \citep{li2020federated} also utilizes weight averaging to perform model aggregation, but with added regularization to tackle the data heterogeneity issue. 

\textbf{Knowledge distillation in federated learning.} KD \citep{hinton2015distilling} has been proposed in deep learning to compress deep neural networks. Student models, which typically have a smaller size, are forced to mimic the output of the teacher model. In some recent works, KD has been applied in FL with two lines of approaches. The first utilizes distillation to perform model aggregation. For example, FedDF \citep{lin2020ensemble} and FedBE \citep{chen2020fedbe} utilized client models as the teacher model to update the global model on the server, and the purpose is to improve the model aggregation performance with heterogeneous data. Some other works, including FedFTG \citep{zhang2022fine} and Fed-ET \citep{cho2022heterogeneous}, utilized distillation for the same purpose, but in a different setting, i.e. in a data-free or model heterogeneous environment. The second approach shares model predictions between clients and the server for training purposes. For example, in FD \citep{jeong2018communication}, model predictions were shared between clients to regularize the local training. In FedAD \citep{gong2021ensemble}, model outputs of the clients were used to train a global model on the server through distillation.

In this paper, we focus on distillation-based model aggregation for FL. The proposed FedSDD utilizes distillation to perform model aggregation, and can be classified into the first approach. In particular, FedSDD first performs local training and weight averaging within multiple groups clients to obtain multiple global models. Then, it builds an ensemble as the teacher model and enhance one of the global model via distillation. The key innovations lie in the enhanced scalability due to the way of building of ensemble and the improved diversity across the compositing models of the ensemble by the designated KD to only one of the global models. Furthermore, the process in the first step is fully modularized, such that averaging based methods like FedAvg \citep{mcmahan2017communication} and FedProx \citep{li2020federated} can be directly applied. As a result, the proposed framework can easily adopt existing FL algorithms for further enhancement.
%-------------------------------------------------------------------------

\section{Scalable and diversity-enhanced distillation for model aggregation}
\subsection{Overview}

\begin{figure}
    \centering
    \scriptsize
    \includegraphics[width=\textwidth]{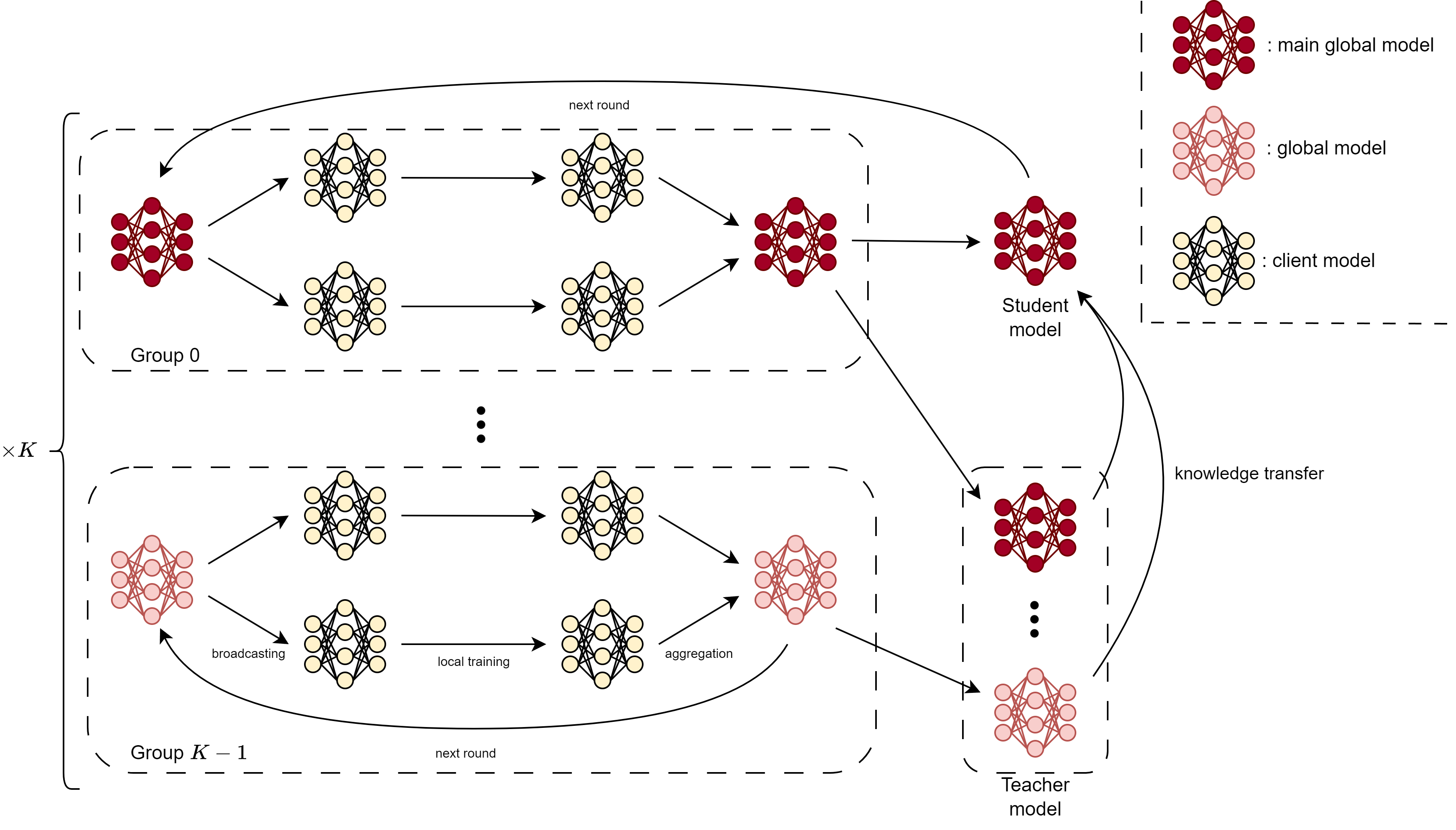}
    \caption{The workflow of FedSSD (w/o temporal ensembling). In FedSSD, there are $K$ global models, where one of which is the main global model and we optimize the main global model thought both local training and knowledge distillation. In each round, all global models will be distributed to different groups of clients for performing local training, and the updated weights from the clients will be aggregated into the new weights of the corresponding global models. Then, these updated global models will form an ensemble, which will be utilized as the teacher model in knowledge distillation, for enhancing the student model, i.e. the main global model. Note that the temporal ensembling is skipped here for ease of presentation.}
    \label{fig:fedssd}
\end{figure}
For ease of illustration, we assume the same training setting as FedDF \cite{lin2020ensemble}, where the labeled data are distributed over a set of clients, in a Non-IID manner, and the server only has access to a small unlabeled dataset. The training target is to learn a global model through multiple rounds of optimization. 
Note that the key techniques of the proposed method can be extended to different settings, e.g. with heterogeneous devices \cite{lin2020ensemble, cho2022heterogeneous}.

In FedSDD, we perform $T$ rounds of training and, in each round, $K$ global models will be trained by $K$ groups of clients. The detailed grouping and training procedure will be introduced later. For simplicity, we assume that these global models share the same architecture, but a distinct set of weights where the initial weights of the $k$-th global model is denoted by $w^{*}_{0, k}$ for $0 \leq k <K$. In the $t$-th round, all $K$ global models are updated and the new weights are denoted by $w^{*}_{t, k}$ for the $k$-th model. The first global model, which 
has weights $w^{*}_{t, 0}$, is also referred to as the \textit{main global model} in FedSDD, and our target is to maximize the performance of this global model. There are two major operations in each round of the proposed algorithm: (i). local training and weight averaging, and (ii). knowledge distillation. The workflow of FedSSD is shown in Fig. \ref{fig:fedssd}. In the following, we introduce the details of the two operations and highlight their difference from existing methods.

\subsubsection{Local training and weight averaging}
\label{subsec:local_training}
Assume there are in total $N_t$ participating clients in the $t$-th round with $1 \leq t < T$. These clients are randomly but evenly distributed into $K$ groups, where $N_{t, k}$ denotes the number of clients in the $k$-th group. Then, the server broadcasts the weights of the $k$-th global model from last round, $w^{*}_{t - 1, k}$, to the $k$-th group of clients to initialize their local models. Thus, the initial model weights for the $n$-th client in the $k$-th group is given by:
\begin{equation}
w^{l}_{t, k, n} = w^{*}_{t - 1, k},
\end{equation}
where $0 \leq n < N_{t, k}$. After that, all clients perform local training with their private datasets $X_{t, k, n}$. The updated weights, denoted by $w^{l*}_{t, k, n}$, will then be collected by the server for weight averaging \cite{mcmahan2017communication} within each group to obtain $K$ averaged models. In particular, the averaged model of the $k$-th group is given by:

\begin{equation}
w_{t, k} = \sum^{N_{t, k} - 1}_{i=0} \frac{|X_{t, k, i}|}{\sum^{N_{t, k} - 1}_{j=0} |X_{t, k, j}|} w^{l*}_{t, k, i}.
\end{equation}

It is worth noticing that, the above workflow performs weights broadcasting, local training, and aggregation in different groups, independently. The workflow within one group is essentially the same as FedAvg \cite{mcmahan2017communication}, and shares similar procedures as many common FL algorithms. Thus, the actual implementation can be easily replaced by other federated learning algorithms, to bring additional benefits, e.g. replacing FedAvg by FedProx \cite{li2020federated} to tackle the Non-IID issue by introducing regularization. Although FedDF \cite{lin2020ensemble} has demonstrated a similar compatibility, FedSDD further improves it by the full modularization of the weight averaging and local training step. As a result, the direct access to the client models are not required by the server. For example, by utilizing secure aggregation \cite{bonawitz2017practical}, the server can obtain the results of aggregation but without accessing the individual client models, which provide better privacy protection to the clients. In this paper, our focus is not on the local training and weight averaging step, thus we use FedAvg \cite{mcmahan2017communication} as the default option. In the experiments, we will demonstrate that FedSDD works well with different alternative FL algorithms. 

\begin{algorithm}[t]
\caption{FedSDD}
\label{alg:fedeed}
\begin{algorithmic}
    \State $T$: Number of training rounds
    \State $N_{t, k}$: Number of participating clients in the $k$-th group at the $t$-th round.
    \State $C_{t, k, n}$: The $n$-th client in the $k$-th group at the $t$-th round.
    \State $K$: Number of global models, $R$: Number of checkpoints for building ensemble.
    \State \hrulefill
    \State  // initialize $K$ global models $w^{*}_{0, 0}, ..., w^{*}_{0, K-1}$
    \For{$t \in \{1, ..., T\}$}
        \For{$k \in \{0, ..., K - 1\}$}\textbf{ in parallel}
            \For{$n \in \{0, ..., N_{t, k} - 1\}$}\textbf{ in parallel}
                \State $w^{l}_{t, k, n} \gets w^{*}_{t-1, k}$
                \State $w^{l*}_{t, k, n} \gets \textit{LocalTraining}(C_{t, k, n}, w^{l}_{t, k, n})$                
            \EndFor
            \State $w_{t, k} \gets \textit{Aggregation}(w^{l*}_{t, k, 0}, ..., w^{l*}_{t, k, N_{t, k} - 1})$
        \EndFor
        \For{$k \in \{0, ..., K - 1\}$}\textbf{ in parallel}
            \If{$k == 0$}
                \State $w^{*}_{t, k} \gets \textit{Distillation}(w_{t, 0}, \{w_{t, 0}, ..., w_{t, K - 1}, ..., w_{t - R + 1, 0}, ..., w_{t - R + 1, K - 1}\})$
            \Else
                \State $w^{*}_{t, k} \gets w_{t, k}$
            \EndIf
        \EndFor
    \EndFor
\end{algorithmic}
\end{algorithm}

\subsubsection{Diversity-enhanced knowledge distillation}
\label{subsec:distillation}
After weight averaging, the server builds an ensemble with the $K$ global models, and utilizes it to perform distillation to enhance the performance of the main global model (i.e. the model with weight $w_{t, 0}$). Note that only the weight of the main global model will be updated in this stage. The purpose is to prevent different global models from learning similar knowledge and thus enhance the diversity. This is expected to generate an ensemble that can outperform individual global models. Let $F(x|w)$ denote the logit output of the model with weight $w$ and input $x$, and denote $\sigma(.)$ as the softmax function. Then, the output of the ensemble $F_{ensemble}$ can be determined by the outputs of $K$ group models as:

\begin{equation}
    y = \sigma\left(F_{ensemble}(x | w_{0, t}, ..., w_{K - 1, t})\right) = \sigma\left(\sum^{K - 1}_{k = 0} \frac{1}{K} F(x | w_{k, t})\right).
\end{equation}

For classification tasks, the final model weights at round $t$ can be determined as:

\begin{equation}
\begin{cases}
    w^{*}_{t, k} = \operatorname*{argmin}_{w} E_{x}\left[KL(\sigma(F(x|w)), \sigma(F_{ensemble}(x|w_{0, t}, ..., w_{K - 1, t})))\right], & \text{if } k = 0\\
    w^{*}_{t, k} = w_{t, k}, & \text{otherwise}
\end{cases}
\end{equation}

where $KL$ denotes the KL-divergence loss \cite{hinton2015distilling, li2020federated}, and the expectation is empirically minimized.

\textbf{Remark 1:} In FedSDD, KD is only performed for the main global model to avoid forcing all clients to learn the same knowledge, i.e., to enhance the diversity. It is also worth mentioning that the members of all groups will be resampled and reshuffled at the beginning of each round. As a result, every global model has a chance to learn from all client datasets.

\textbf{Remark 2:} The size of the ensemble is determined by the number of groups, $K$, instead of the number of participating clients, $N_t$. As a result, the computational cost of KD in FedSDD will not increase with  the number of participating clients, offering a better scalability than exiting methods.

\subsubsection{Temporal ensembling}
\label{subsec:temp_ensemling}
Although a large number of global models (i.e. a large $K$) may enhance the ensemble capability, it may also slow down the convergence of individual global models.
To enhance the ensemble capability without sacrificing the convergence of individual global models, we propose to build the ensemble by leveraging multiple checkpoints of different training rounds. Intuitively, global models from consecutive rounds were trained based on the data from different subsets of the clients. Thus, ensembling global models from multiple checkpoints of different rounds can emulate the case where more clients are participating. This allows us to build a larger ensemble without slowing down individual models' convergence.

With temporal ensembling, the output of the ensemble can be determined as:

\begin{equation}
    y = \sigma(F_{ensemble}(x | w_{0, t}, ..., w_{K - 1, t - R + 1}))
    = \sigma\left(\sum^{K - 1}_{k = 0} \sum^{R - 1}_{r = 0} \frac{1}{KR} F(x | w_{k, t - r})\right)
\end{equation}

where $R$ denotes the number of rounds utilized for building the ensemble. In our experiments, we found that building the ensemble from multiple rounds is highly effective, especially when the degree of Non-IID is high.

By combining the above techniques, the FedSDD algorithm is shown Algo. \ref{alg:fedeed} whose work flow is given in Fig. \ref{fig:fedssd}. 

\subsection{Comparison with existing works}
In this section, we compare FedSDD with existing methods. 

\label{subsec:comparison}
\begin{table}[t]
	\centering
	\scriptsize
	\caption{Comparison between different distillation-based methods. }
        \begin{tabular}[t]{ccccc}
		\toprule
		Method & Ensemble size & Parallelism between server/clients & No access to client models \\
		\midrule[1pt]
		FedDF & $C$ & \ding{55} &  \ding{55} \\
		FedBE & $C+S+1$ & \ding{55} & \ding{55} \\
		FedSDD (w/ FedAvg) & $KR$ &\ding{51} & \ding{51} \\
		\midrule
		\multicolumn{5}{c}{$K$: Number of global models	\quad$R$: Number of checkpoint rounds	\quad $C$: Number of active clients \quad $S$: Number of sampled models} \\
		\bottomrule
	\end{tabular}
	%\hfill
	\label{tab:methods_compare}
\end{table}

\textbf{Distillation-based aggregation.} Although existing works \citep{lin2020ensemble, zhang2022fine, huang2022learn, cho2022heterogeneous} have already utilised distillation to enhance the global model, they typically did not concern about the scalability of the methods. For example, with FedDF \cite{lin2020ensemble}, the cost of performing inference with the ensemble will be high especially for systems with a large number of clients, because the ensemble output is the average of that from all client models.

As mentioned in Remark 2, FedSDD trains multiple global models and builds the ensemble from them directly. By training $K$ global models and storing $R$ checkpoints for each of them, the ensemble is built by $K \times R$ models. 
As a result, the complexity of the ensemble can be controlled by adjusting $K$ and $R$, instead of scaling with the number of clients.

\textbf{Multiple global models.} Multiple models have been utilized for both inference and learning purposes. For example, Fed-ensemble \cite{shi2021fed} learns multiple models and build the ensemble for inference. However, if the ensemble is large, it can not be directly utilized by clients due to their limited computation resources. The heterogeneous version of FedDF \cite{lin2020ensemble} trains multiple global models with different architectures and scales, and performs distillation to enhance all of them, which may ruin the diversity among models.

Different from existing works, FedSSD (i). targets on training a single, compact, and high performance global model; and (ii). uses KD to update only the main global model, which preserves the diversity and achieves enhanced performance.

\textbf{Diversity preserving knowledge distillation.} Prior works in online KD (i.e., not for FL) have proposed different ways in preserving the models' diversity during training. Codistillation \cite{anil2018large} applied the distillation term in its loss function, after the training has been started for a while. OKDDip \citep{chen2020online} applied a similar scheme with one main student for distillation, but still utilized KD to enhance the remaining student models. Comparing with OKDDip, the KD scheme in this paper is not only simpler with reduced computation complexity (no forward/backward pass for the remaining $K-1$ global models), but also allows parallel training between the server-side and the client-side which can benefit FL. See supplementary materials for more details.  

\textbf{Temporal ensembling.} Prior works utilized multiple checkpoints to build ensembles for prediction \cite{huangsnapshot} or semi-supervised learning \cite{lainetemporal}. These works either trained models for a number of epochs with learning rate scheduling to obtain a diverse set of checkpoints at different local minimas, or utilized data augmentation and model regularization to produce predictions with the ensemble. FedGKD \cite{yao2021local} utilized historical checkpoints for regularizing the training of client models. In this paper, temporal ensembling is utilized in FL systems for enhancing the capability of the teacher model. Moreover, temporal ensembling in FedSDD can be simply achieved by combining model checkpoints of consecutive rounds without any additional efforts. This is possible due to the Non-IID learning environment in FL, and can prevent the slowing down of convergence that may occur with the above-mentioned techniques.

\textbf{Additional advantage of FedSDD.} FedSDD also provides additional benefits beyond scalability and performance. Unlike previous works, FedSDD fully modularizes the local training and the weight averaging step, and does not require direct access to the client models. As a result, FedSDD can obtain the same privacy protection level as weight averaging methods such as FedAvg \cite{mcmahan2017communication}. For example, secure aggregation \cite{bonawitz2017practical} can be used to prevent the server from obtaining sensitive information from individual client models, which can't be achieved by existing knowledge distillation based methods \cite{wang2021field}. Furthermore, in FedSDD, only one global model, i.e. the main global model, is updated by KD, thus the training of the other $K - 1$ global models can start immediately, after the previous round of training is completed. This allows for parallelism between the server-side distillation and the client-side local training. The parallelism can significantly improve the efficiency of federated learning. 

We summarise the advantages of FedSDD compared with FedDF \cite{lin2020ensemble} and FedBE \cite{chen2020fedbe} in Table \ref{tab:methods_compare}.

%-------------------------------------------------------------------------

\section{Experiments}
\label{sec:experiement}

\subsection{Image Classification with CIFAR10/100.}
\begin{table}[h]
	\centering
	\scriptsize
	%{\tabcolsep}{2pt}
	\caption{CIFAR10/100 results.}
	\begin{tabular}[t]{@{} c c c c c c }
		\toprule
		&  & \multicolumn{2}{c|}{CIFAR10} & \multicolumn{2}{c}{CIFAR100} \\
		Model & Method & $\alpha = 1.0$ & $\alpha = 0.1$ & $\alpha = 1.0$ & $\alpha = 0.1$  \\
		\midrule[1pt]
		ResNet20 & FedAvg & $88.53 \pm 0.31$ & $78.72 \pm 2.31$ & $58.84 \pm 0.42$ & $52.98 \pm 1.41$ \\
		& FedProx & $88.36 \pm 0.18$ & $79.44 \pm 2.17$& $58.76 \pm 0.93$ & $53.57 \pm 0.37$ \\
		& SCAFFOLD & \underline{$89.85 \pm 0.50$} & $80.08 \pm 1.53$ & $61.09 \pm 0.46$ & $54.74 \pm 1.26$ \\
		& FedDF & $87.98 \pm 0.16$ & $80.04 \pm 1.87$ & $59.74 \pm 0.38$ & $51.83 \pm 0.83$ \\
		& FedSDD ($R=1$) & $89.06 \pm 0.19$ & $80.18 \pm 2.38$ & $61.90 \pm 0.83$ & $54.72 \pm 0.87$ \\
		& FedSDD ($R=2$) & $89.01 \pm 0.18$ & $81.27 \pm 2.16$ & $62.42 \pm 0.57$ & $56.39 \pm 1.10$ \\
		& FedSDD ($R=4$) & $89.01 \pm 0.07$ &  \underline{$82.21 \pm 1.10$} & \underline{$62.84 \pm 0.45$} & \underline{$56.90 \pm 0.77$} \\
		\midrule[1pt]
		ResNet56 & FedAvg & $89.25 \pm 0.11$ & $80.05 \pm 2.65$ & $59.47 \pm 1.23$ & $54.37 \pm 1.24$ \\
		& FedProx & $89.65 \pm 0.23$ & $80.84 \pm 1.32$ & $61.01 \pm 1.02$ & $57.18 \pm 0.92$ \\
		& SCAFFOLD & \underline{$90.73 \pm 0.27$} & $82.22 \pm 0.92$ & $62.94 \pm 0.91$ & $56.36 \pm 0.31$ \\
		& FedDF & $88.87 \pm 0.29$ & $81.10 \pm 2.31$ & $60.71 \pm 0.68$ & $52.52 \pm 1.37$ \\
		& FedSDD ($R=1$) & $89.86 \pm 0.14$ & $81.30 \pm 2.17$ & $62.95 \pm 2.43$ & $57.48 \pm 2.03$ \\
		& FedSDD ($R=2$) & $90.23 \pm 0.23$ & $82.77 \pm 1.67$ & $65.28 \pm 0.64$ & $59.37 \pm 0.64$ \\
		& FedSDD ($R=4$) & $90.00 \pm 0.26$ & \underline{$83.20 \pm 0.99$} & \underline{$65.63 \pm 0.89$} & \underline{$59.56 \pm 0.43$} \\
		\midrule[1pt]
		WRN16-2 & FedAvg & $90.26 \pm 0.33$ & $81.52 \pm 2.66$ & $63.57 \pm 1.06$ & $58.77 \pm 0.55$ \\
		& FedProx & $90.38 \pm 0.29$ & $81.78 \pm 2.58$ & $63.75 \pm 0.84$ & $59.24 \pm 1.00$ \\
		& SCAFFOLD& \underline{$91.32 \pm 0.11$} & $82.90 \pm 2.19$ & $65.33 \pm 0.48$ & $60.66 \pm 0.85$ \\
		& FedDF & $89.52 \pm 0.10$ & $81.32 \pm 2.91$ & $63.54 \pm 0.53$ & $56.89 \pm 1.28$ \\
		& FedSDD ($R=1$) & $90.69 \pm 0.41$ & $82.62 \pm 2.46$ & $66.46 \pm 0.30$ & $61.16 \pm 1.20$ \\
		& FedSDD ($R=2$) & $90.77 \pm 0.16$ & $83.39 \pm 2.17$ & $67.11 \pm 0.41$ & $62.37 \pm 0.84$ \\
		& FedSDD ($R=4$) & $90.70 \pm 0.31$ & \underline{$84.02 \pm 1.75$} & \underline{$67.48 \pm 0.22$} & \underline{$63.05 \pm 0.80$} \\
		\bottomrule
	\end{tabular}
	%\hfill
	\label{tab:cifar_results}
\end{table}

\begin{table}[t]
	\centering
	\scriptsize
	\caption{Round time comparison between FedAvg\cite{mcmahan2017communication}, FedDF\cite{lin2020ensemble} and FedSSD. Round times of FedDF and FedSDD are reported by the relative time to FedAvg, i.e. the time spent on KD. All results are in seconds.}
        \begin{tabular}[t]{ccccc}
		\toprule
		& \multicolumn{3}{c}{Number of clients} \\
		Method & 8 & 14 & 20 \\
	\midrule[1pt]
		FedAvg & 123.6 &  214.9 & 301.6 \\
		FedDF & +223.6 & +337.0 & +447.1 \\
		FedSDD & +174.1 & +177.1 & +175.5 \\
		%FedAvg & 122.9 & 213.9 & 304.0 \\
		%FedDF & +175.6 & +246.9 & +315.6 \\
		%FedSDD & +175.0 & +173.4 & +178.9 \\
  \bottomrule
	\end{tabular}
	%\hfill
	\label{tab:speeds_compare}
\end{table}

In the experiments, we will perform classification tasks following the same setting as \cite{lin2020ensemble}. In particular, the clients have the CIFAR10/100 \citep{krizhevsky2009learning} datasets for local training. Similar to other FL algorithms with distillation (i.e. FedDF \cite{lin2020ensemble} and FedSDD), the server holds unlabelled data from the CIFAR100/ImageNet (resized to $32\times32$) \citep{deng2009imagenet} to perform distillation for classifying images from the CIFAR10/100 datasets, respectively. There are in total 20 clients, where 40\% of them participate in the training in each round, with 100 rounds of training in total. We sampled data for each client by the Dirichlet distribution \citep{hsu2019measuring}, with $\alpha = \{1.0, 0.1\}$ for different levels of data heterogeneity. For each setting, we reported the average top-1 accuracy of three runs with different seeds, where the same seeds were shared across different methods to generate the same data partition and sequence of clients' participation for a fair comparison.
We trained ResNet20/56 \citep{he2016deep} and WRN16-2 \citep{zagoruyko2016wide}, which have been proven capable to achieve high accuracy on CIFAR10/100 datasets. For the optimizer, we used SGD with learning rate of 0.8 and 0.1 and batch size of 64 and 256, for the client-side and the server-side training, respectively. We did not apply weight decay or learning rate scheduling in our experiments.

We compared FedSDD with FedAvg \citep{mcmahan2017communication}, FedProx\citep{li2020federated}, SCAFFOLD\citep{karimireddy2020scaffold} and FedDF\citep{lin2020ensemble}. We set the number of local training epochs to be 40, and the regularizer parameter $\mu$ of FedProx to be 0.001. For FedSDD, we set the number of global models $K$ to be 4 and the number of checkpoints $R$ to be 1/2/4 to build the ensemble. For both FedDF and FedSDD, we performed distillation with 5000 steps in each training round, with the temperature $\tau$ to be 4.0.

The results are shown in Table \ref{tab:cifar_results}. It can be observed that FedSDD outperformed all other methods on the harder dataset, CIFAR100, for all settings with high degree of data heterogeneity. It suggests that FedSDD is a good candidate for tackling the Non-IID issue. We also found that the value of $R$ has little impact to the performance of FedSDD in the CIFAR10 experiments with low degree of data heterogeneity, but a higher value of $R$ improved the performance of FedSDD significantly in the CIFAR10 runs with more Non-IID data, and on CIFAR100 datasets with both high and low degrees of data heterogeneity. Note that in the above experiments, FedSDD with $R \leq 2$ has an ensemble size no larger than that of FedDF, thereby maintaining an overall lower computational cost. In such cases, FedSDD can still outperform FedDF.

%Moreover, FedSDD has lower per-round communication cost  than FedProx and SCAFFOLD, since it only require sending global models/client models, where FedProx and SCAFFOLD send additional data for providing regularaztion to tackling the data heterogeneity issue (the same size as the model).

Next, we illustrate the advantage of FedSSD in terms of scalability. Due to the lack of access to large scale FL system, we can only provide the comparison with small scale experiments. Table \ref{tab:speeds_compare} shows the comparison of the round time (execution time per communication round) between FedAvg\cite{mcmahan2017communication}, FedDF\cite{lin2020ensemble} and FedSSD. The experiment is conduced on an A100 GPU, with the setting of the CIFAR10/100 experiment, where the model is ResNet-20. It can be observed that, with 8 clients, the time spent by FedSDD on KD is around 78\% of that for FedDF. However, since the ensemble size depends on only the number of global models ($K=4$ for this experiment), when increasing the number of clients to 14/20, the time spent on KD reduces to ~52\%/39\% of FedDf. This demonstrates the scalability advantage of FedSDD, and the scalability is essential for applying KD in large scale FL systems.

%-------------------------------------------------------------------------

\section{Ablation Studies}
\label{sec:ablation}
\subsection{Compatibility of FedSDD with other FL algorithms}
\begin{table}[t]
	\centering
	\scriptsize
	%{\tabcolsep}{2pt}
	\caption{CIFAR10 results of FedSDD with different local training settings. }
	\begin{tabular}[t]{@{} c c | c c | c c }
		\toprule
		&  & \multicolumn{2}{c|}{CIFAR10 (8 clients per round)} & \multicolumn{2}{c}{CIFAR10 (20 clients per round)} \\
		Model & Method & $\alpha = 1.0$ & $\alpha = 0.1$ & $\alpha = 1.0$ & $\alpha = 0.1$  \\
		\midrule[1pt]
		ResNet20 & FedAvg & $88.53 \pm 0.31$ & $78.72 \pm 2.31$ & $88.59 \pm 0.28$ & $79.84 \pm 1.50$ \\
		& FedProx & $88.36 \pm 0.18$ & $79.44 \pm 2.17$& $88.52 \pm 0.07$ & $79.63 \pm 2.48$ \\
		& SCAFFOLD & \underline{$89.85 \pm 0.50$} & $80.08 \pm 1.53$ & $90.10 \pm 0.07$ & $82.83 \pm 2.40$ \\
		& FedSDD w/ FedAvg & $89.06 \pm 0.19$ & $80.18 \pm 2.38$ & $89.04 \pm 0.11$ & $82.89 \pm 1.30$ \\
		& FedSDD w/ FedProx & $88.98 \pm 0.22$ & \underline{$81.21 \pm 1.30$} & $89.28 \pm 0.17$ & $82.12 \pm 2.26$ \\
		& FedSDD w/ SCAFFOLD & $88.80 \pm 0.16$ & $76.51 \pm 3.74$ & \underline{$90.66 \pm 0.04$} & \underline{$84.30 \pm 1.88$} \\
		\bottomrule
	\end{tabular}
	%\hfill
	\label{tab:localtrainers_results}
\end{table}

As mentioned in section \ref{subsec:local_training}, the local training and model aggregation step of FedSDD can be replaced by many different existing methods. We evaluate the compatibility of FedSDD with different choices of local training scheme by combining FedSDD with FedAvg \citep{mcmahan2017communication} (default option in this paper), FedProx \citep{li2020federated}, and SCAFFOLD \citep{karimireddy2020scaffold}.
The results in Table \ref{tab:localtrainers_results} (left, with 8 clients) show that FedSDD is compatible with FedProx and the combination achieves better performance with high degree of data heterogeneity.

However, the combination of FedSDD and SCAFFOLD did not improve the performance under the concerned setting. We conjectured this may be due to the fact that the number of clients per global model is only 2 in FedSDD which is too small for SCAFFOLD. We thus performed additional experiments with 20 clients per round, i.e. the number of clients per global models became 5, as shown in Table \ref{tab:localtrainers_results} (right). With these settings, the combination of FedSDD and SCAFFOLD achieved much higher accuracy than the stand-alone FedSDD and SCAFFOLD. The above results suggest that FedSDD is flexible with the implementation of the local training scheme.

\subsection{Comparison between different ways of building the ensemble}
\begin{table}[t]
	\centering
	\scriptsize
	%{\tabcolsep}{2pt}
	\caption{CIFAR10 results of different ensemble settings.}
	\begin{tabular}[t]{@{} c c | c c }
		\toprule
		&  & \multicolumn{2}{c}{CIFAR10}\\
		Model & Method & $\alpha = 1.0$ & $\alpha = 0.1$ \\
		\midrule[1pt]
		ResNet20 & Global model ($K=1$) & $88.53 \pm 0.31$ & $78.72 \pm 2.31$  \\
		& Ensemble ($K=1$) Clients) & $88.82 \pm 0.21$ & $80.22 \pm 0.88$ \\
		& Ensemble ($K=1$, Bayesian, Gaussian) & $88.72 \pm 0.14$ &  $79.60 \pm 1.26$ \\
		& Ensemble ($K=1$, Bayesian, Dirichlet) & $88.70 \pm 0.11$ &  $80.25 \pm 1.43$ \\
		\cmidrule(lr){2-4}
		& Global model ($K=4$)  & $86.69 \pm 0.54$ & $70.74 \pm 5.11$ \\
		& Ensemble ($K=4$, Clients) & $90.49 \pm 0.17$ & $82.12 \pm 1.65$ \\
		& Ensemble ($K=4, R=1$, Aggregated) & $90.58 \pm 0.31$ & $81.76 \pm 2.40$ \\
		& Ensemble ($K=4, R=2$, Aggregated) & \underline{$90.75 \pm 0.24$} & $83.50 \pm 1.56$ \\
		& Ensemble ($K=4, R=4$, Aggregated) & $90.69 \pm 0.19$ & \underline{$83.99 \pm 0.88$} \\
		\bottomrule
	\end{tabular}
	%\hfill
	\label{tab:ensemble_cifar_results}
\end{table}

We performed experiments to compare different ways of building the ensemble from global models, including the methods that use client models \cite{lin2020ensemble, chen2020fedbe} and the proposed method which utilizes multiple global models. Following the notations of FedSDD, we use $K$ to denote the number of global models and $R$ to represent the number of checkpoints in building the ensemble. In this experiment, we set $K \in \{1, 4\}$ and $R\in \{1, 2, 4\}$. Note that, to avoid the possible negative impact to the ensembles, we did not use distillation in this experiment.

We considered two existing methods that used only one global model ($K = 1$): (i). Following FedDF \citep{lin2020ensemble}, we utilized all client models to build the ensemble and only trained one global model; (ii). Following FedBE \citep{chen2020fedbe}, we trained a global model and built the ensemble by combining the client models, the averaged model, and $10$ models sampled from the Gaussian distribution or the Dirichlet distribution. 

Then, we considered two strategies which utilize multiple global models ($K > 1$): (iii). Similar to (i), we utilized client models (initialized from different global models) to build the ensemble. (iv). Similar to FedSDD, we built the ensemble from multiple checkpoints of global models, which is also the case for Fed-ensemble when $R=1$.

The results are shown in Table \ref{tab:ensemble_cifar_results}, where we also provide the performance of the standalone global model for reference purposes. With low degree of heterogeneity ($\alpha=1.0$), despite slowing down the convergence of individual global models, a larger number of global models offers better results. All ensembles with multiple global models performed closely, where the ensembles built by global models from multiple checkpoints achieved slightly higher accuracy. With $K=1$, all methods did not show clear performance advantage over the global model, which agrees with our understanding about the diversity among the compositing models.

With highly Non-IID data, all ensemble strategies showed significant improvement over the baseline global model with $K=1$. However, strategies with multiple global models still outperformed the methods with single global model, and the best results were obtained by strategy (iv) with $R=4$. It indicates that direct access to client models is not necessary to build a high performance ensemble.

In the above experiments, ensembles built by the multiple checkpoints of the global models achieved the best performance, which is the strategy being used in FedSDD.

\subsection{Comparison between different ways of distillation}
\begin{table}[t]
	\centering
	\scriptsize
	%{\tabcolsep}{2pt}
	\caption{CIFAR10 results of different distillation settings.}
	\begin{tabular}[t]{@{} c c | c c}
		\toprule
		&  & \multicolumn{2}{c}{CIFAR10} \\
		Model & Method & $\alpha = 1.0$ & $\alpha = 0.1$ \\
		\midrule[1pt]
		ResNet20 & w/o distillation & $86.69 \pm 0.54$ & $70.74 \pm 5.11$ \\
		& Basic distillation & $88.63 \pm 0.34$ & $79.98 \pm 2.42$ \\
		& Basic distillation w/ warm-up (20 rounds) & $88.66 \pm 0.22$ &  \underline{$80.20 \pm 1.81$} \\
		& Basic distillation w/ warm-up (40 rounds) & $88.47 \pm 0.04$ & $79.27 \pm 2.14$ \\
		& Diversity preserved distillation & \underline{$89.06 \pm 0.19$} & $80.18 \pm 2.38$ \\
		\midrule[1pt]
		ResNet20 (Ensemble) & w/o distillation & \underline{$90.58 \pm 0.31$} & \underline{$81.76 \pm 2.40$} \\
		& Basic distillation & $89.23 \pm 0.36$ & $80.57 \pm 2.24$ \\
		& Basic distillation w/ warm-up (20 rounds) & $89.38 \pm 0.42$ & $80.80 \pm 1.95$ \\
		& Basic distillation w/ warm-up (40 rounds) & $89.19 \pm 0.24$ & $80.32 \pm 2.55$ \\
		& Diversity preserving distillation & $90.31 \pm 0.20$ & $81.39 \pm 2.54$ \\
		\midrule[1pt]
	\end{tabular}
	%\hfill
	\label{tab:distill_cifar_results}
\end{table}

In FedSDD, distillation is performed to enhance only the main global model, which enhances the diversity. To evaluate the effectiveness of the proposed method, we performed comparison between different ways of distillation: (i). A basic KD scheme, which utilizes the ensemble as the teacher model and all global models are trained by mimicking the output of the ensemble. Note that we fixed the weights of the ensemble during the distillation process. This is similar to the heterogeneous version of FedDF \citep{lin2020ensemble}, where all global models are improved by KD; (ii). A scheme following Codistillation \citep{anil2018large}, which skips the distillation in the early rounds of training, such that the models will not learn from each other at the beginning to maintain potentially higher diversity. We set the number of warm-up epochs to 20/40; and (iii). The proposed diversity-enhanced distillation, where only the main global model is enhanced by KD. 

Results in Table \ref{tab:distill_cifar_results} show that, the global model trained by the proposed scheme outperformed its counterparts with low degree of data heterogeneity, and performed similarly as the others with high degree of data heterogeneity. In fact, the ensemble trained by the proposed scheme obtained the best result for all the cases, which is close to the ensemble without distillation. These observations support our expectation that diversity enhancing schemes can  improve the performance of both the ensemble and the distilled global model.

\section{Conclusion}
In this paper, we proposed FedSSD, a distillation-based federated learning algorithm to tackle the limitation of existing KD-based FL algorithms for model aggregation. The proposed scheme is highly scalable as it utilizes multiple global models, instead of all client models, for building the ensemble. The ensemble in FedSSD has a better performance due to the diversity-enhancing KD scheme, which in turn improves the performance of the global model through KD. Furthermore, the main modules of FedSSD are compatible with different FL algorithms, whose adaptation can further improve the performance of FL systems.

%-------------------------------------------------------------------------

{
\small
\bibliographystyle{abbrv}
\bibliography{egbib}
}

%%%%%%%%%%%%%%%%%%%%%%%%%%%%%%%%%%%%%%%%%%%%%%%%%%%%%%%%%%%%

\newpage

\appendix

\section{Appendix}

% \subsection{Additional figures for comparing different ensemble strategies}

% (To be updated)
% In this section, we provide the plots of the experiment in section \ref{subsec:ensembletrainer_exp}, as shown in Fig. \ref{fig:ensemble_cifar10}. The settings with 4 global models (i.e. $K=4$) consistently outperform the others, where the scheme utilizing global models as the ensemble performs closely as that utilizing client models. Utilizing multiple checkpoints of global models as the ensemble obtained the best result in the experiment.

% \begin{figure}
% 	\centering
% 	\begin{subfigure}[b]{0.495\textwidth}
% 		\centering
% 		\includegraphics[width=\textwidth]{fig/ensemble_low.png}
% 		\caption{$\alpha = 1.0$}
% 		\label{fig:ensemble_cifar10_low}
% 	\end{subfigure}
% 	\begin{subfigure}[b]{0.495\textwidth}
% 		\centering
% 		\includegraphics[width=\textwidth]{fig/ensemble_high.png}
% 		\caption{$\alpha = 0.1$}
% 		\label{fig:ensemble_cifar10_high}
% 	\end{subfigure}
% 	\caption{Comparison between different ensemble strategies with CIFAR-10. Note that we applied smoothing to the curves to improve the visibility.}
% 	\label{fig:ensemble_cifar10}
% \end{figure}

\subsection{Experiment with different communication intervals}

\begin{table}[ht]
	\centering
	\scriptsize
	%{\tabcolsep}{2pt}
	\caption{CIFAR10 results with different communication intervals}
	\begin{tabular}[t]{@{} c c c | c c}
		\toprule
		& &  & \multicolumn{2}{c}{CIFAR10} \\
		\# rounds / \# local epochs & Model & Method & $\alpha = 1.0$ & $\alpha = 0.1$ \\
		\midrule[1pt]
		25 / 160 & ResNet20 & FedAvg & $85.81 \pm 0.12$ & $73.17 \pm 3.66$ \\
		&  & FedDF & \underline{$86.78 \pm 0.12$} & \underline{$77.79 \pm 1.73$} \\
		&  & FedSDD & $86.27 \pm 0.68$ & $73.99 \pm 2.15$ \\
		\midrule[1pt]
		100 / 40 & ResNet20 & FedAvg & $88.53 \pm 0.31$ & $78.72 \pm 2.37$ \\
		&  & FedDF & $87.98 \pm 0.16$ & $80.04 \pm 1.87$ \\
		&  & FedSDD & \underline{$89.06 \pm 0.19$} & \underline{$80.18 \pm 2.38$} \\
		\midrule[1pt]
		400 / 10 & ResNet20 & FedAvg & $89.69 \pm 0.27$ & $79.19 \pm 0.42$ \\
		&  & FedDF & $88.90 \pm 0.19$ & $78.92 \pm 0.59$ \\
		&  & FedSDD & \underline{$90.37 \pm 0.08$} & \underline{$81.34 \pm 0.81$} \\
		\midrule[1pt]
	\end{tabular}
	%\hfill
	\label{tab:cifar10_rounds}
\end{table}

We compared the performance of FedSDD with FedAvg \citep{mcmahan2017communication} and FedDF \citep{lin2020ensemble}, with different communication intervals. To fix the total computation workload, we set the number of rounds and the number of local epochs per round as 25/100/400 and 160/40/10, respectively. We also scaled the number of distillation steps to 20000/5000/1250.  From Table \ref{tab:cifar10_rounds}, it can be observed that FedSDD performed better than FedAvg in all the cases, and achieved better results than FedDF in most of the cases. But, with 25 communication rounds, FedDF outperformed FedSDD. We expect this is due to the extremely small number of clients per global models in FedSDD, i.e. two clients per global models, which slowed down the convergence when the communication frequency is too low.

\subsection{Experiment with different number of global models}

\begin{table}[t]
	\centering
	\scriptsize
	%{\tabcolsep}{2pt}
	\caption{CIFAR10 results of FedSDD with different number of global models.}
	\begin{tabular}[t]{@{} c c c | c c}
		\toprule
		& & & \multicolumn{2}{c}{CIFAR10} \\
		Model & Method & $K$ & $\alpha = 1.0$ & $\alpha = 0.1$ \\
		\midrule[1pt]
		ResNet20 & FedSDD & $2$ & $88.34 \pm 0.32$ & $79.38 \pm 2.54$ \\
		& & $3$ & $88.89 \pm 0.22$ & $79.63 \pm 6.15$ \\
		& & $4$ & \underline{$89.06 \pm 0.19$} & \underline{$80.18 \pm 2.38$} \\
		\midrule[1pt]
	\end{tabular}
	%\hfill
	\label{tab:cifar10_num_models}
\end{table}

In the main paper, we used $K=4$ as the default setting of FedSDD. Here, we provide results with $K=2/3/4$. Given there are only 8 clients per round in the default configuration, there are at least two clients per global model, and all global models are aggregated models. For $K=3$, we allocated one more client to the main global model in FedSDD. The results in Table \ref{tab:cifar10_num_models} show that $K=4$ is the best configuration among the tested cases. However, the optimal value of $K$ may vary with different tasks.

\subsection{Experiment with different number of clients}
\begin{table}[t]
	\centering
	\scriptsize
	%{\tabcolsep}{2pt}
	\caption{CIFAR10 results of FedSDD with different scaling schemes.}
	\begin{tabular}[t]{@{} c c c c | c c}
		\toprule
		& & & & \multicolumn{2}{c}{CIFAR10} \\
		\# Clients & Model & Method & $K$ & $\alpha = 1.0$ & $\alpha = 0.1$ \\
		\midrule[1pt]
		8 & ResNet20 & FedAvg & $-$ & $88.53 \pm 0.31$ & $78.72 \pm 2.37$ \\
		&  & FedDF & $-$ & $87.98 \pm 0.16$ & $80.04 \pm 1.87$ \\
		&  & FedSDD & $4$ & \underline{$89.06 \pm 0.19$} & \underline{$80.18 \pm 2.38$} \\
		\midrule[1pt]
		14 & ResNet20 & FedAvg & $-$ & $88.67 \pm 0.22$ & $80.25 \pm 1.64$ \\
		&  & FedDF & $-$ & $88.38 \pm 0.16$ & $81.61 \pm 1.94$ \\
		&  & FedSDD & $4$ & $89.14 \pm 0.09$ & $81.26 \pm 1.99$ \\
		&  & & $7$ & \underline{$89.31 \pm 0.06$} & \underline{$81.90 \pm 1.80$} \\
		\midrule[1pt]
		20 & ResNet20 & FedAvg & $-$ & $88.59 \pm 0.28$ & $79.84 \pm 1.50$ \\
		&  & FedDF & $-$ & $88.12 \pm 0.25$ & $81.55 \pm 2.02$ \\
		&  & FedSDD & $4$ & $89.04 \pm 0.11$ & \underline{$82.89 \pm 1.30$} \\
		&  & & $10$ & \underline{$89.14 \pm 0.11$} & $81.97 \pm 1.54$ \\
		\midrule[1pt]
	\end{tabular}
	%\hfill
	\label{tab:cifar10_scaling}
\end{table}
\begin{table}[ht!]
	\centering
	\scriptsize
	%{\tabcolsep}{2pt}
	\caption{The difference between FedDF and FedSDD. The method in the last row is identical to the FedSDD with $R=1$.}
	\begin{tabular}[t]{@{} c c | c c}
		\toprule
		&  & \multicolumn{2}{c}{CIFAR10} \\
		Model & Method & $\alpha = 1.0$ & $\alpha = 0.1$ \\
		\midrule[1pt]
		ResNet20 & FedDF & $82.53 \pm 0.24$ & $68.72 \pm 4.23$ \\
		& + Improved training configuration & $87.93 \pm 0.47$ & $79.01 \pm 2.15$ \\
		& + Removal of drop-worst \& early-stopping & $87.42 \pm 0.19$ & $78.55 \pm 3.00$ \\
		& + Aggregated ensemble + $K = 4$ + diversity-enhanced distillation& \underline{$88.35 \pm 0.18$} & \underline{$79.59 \pm 2.93$} \\
		\midrule[1pt]
	\end{tabular}
	%\hfill
	\label{tab:fedf_vs_fedeed}
\end{table}

In Table \ref{tab:cifar10_scaling}, we show how the performance of FedSDD scales with the number of clients, and compared the result with that of FedAvg \citep{mcmahan2017communication} and FedDF \citep{lin2020ensemble}. As the number of clients increases, we considered two options: 1.) Scaling the number of clients per global model, i.e. fixing the value of $K$; and 2.) Scaling the number of global models, i.e. fixing the number of clients per group ($K=7$ and $10$ in Table \ref{tab:cifar10_scaling}). We found that FedSDD outperformed FedAvg and FedDF with both scaling schemes, and obtained a better result when the number of global models increases, except the case with high degree of data heterogeneity and 20 clients. In practice, the scaling scheme is subject to the server's capacity, as increasing the number of global models will also increase the cost of distillation.

\subsection{Improvement from FedDF}

Both FedSDD and FedDF \citep{lin2020ensemble} utilize KD for model aggretation with an unlabelled dataset. In this section, we progressively modify FedDF to approach FedSDD, and show the effect of each modification. Note that, following \cite{lin2020ensemble}, we keep 10\% of the training data from CIFAR10 as the validation set of the server in this experiment.

The original FedDF utilized Adam optimizer \citep{kingma2014adam} for server training, and applied the `drop-worst' mechanism and the early-stopping. We performed the following modifications: 1) Adopt the training configuration from the main paper (including the choice of learning rate, the use of SGD optimizer for distillation, etc.); 2) Remove the `drop-worst' and early-stopping operations; and 3) Utilize four aggregated global models to build the ensemble with the proposed distillation scheme, which enhances the model diversity. By introducing the three modifications, the modified FedDF is effectively the same as FedSDD (without temporal ensembling).

Table \ref{tab:fedf_vs_fedeed} shows the results of different modifications, from which we have three observations. First, the configuration we used in this paper improved FedDF. Second, the removal of the `drop-worst' and early-stopping operations led to a worse result, but it saved the need for the validation set at the server. Third, the main mechanisms we used in FedSDD, i.e. the ensemble from aggregated models and the proposed distillation scheme, provided significant performance improvement.

\subsection{Possible extensions of FedSDD}
To perform distillation, a server dataset or generator is mandatory. In this paper, we consider the use of FedSDD with unlabelled server dataset, following \cite{lin2020ensemble, chen2020fedbe}. Since the dataset is not labelled and can be independent of the client datasets, it is cheap to collect. Note that extending FedSDD to include labeled dataset or generator is straightforward. Existing works \cite{lin2020ensemble, zhang2022fine} have utilized generator for knowledge distillation. We can also extend FedSDD to the setting with heterogeneous models \citep{lin2020ensemble,cho2022heterogeneous}, where we can train multiple global models for each model type, and select one main global model for each model type. As long as only global models are utilized to form the ensemble, and KD is only performed for enhancing a subset of the global models, the scalability of the algorithm and the diversity among the global models can be preserved.

% \subsection{Privacy concern of FedSDD}
% \label{subsec:privacy}

% Existing distillation-based model aggregation methods utilized client models as the teacher model for enhancing a global model. For example, FedDF \citep{lin2020ensemble} is built upon weight averaging based methods like FedAvg \citep{mcmahan2017communication}. 

% Weight averaging based methods can benefit from techniques like secure aggregation \citep{bonawitz2017practical}, where individual client updates do not need to be disclosed to the server but the server can still obtain their weighted sum. However, the existing distillation based methods are not compatible with the techniques like secure aggregation \citep{wang2021field}. As a result, the clients are required to send their models or updates to the server, and the server will store the client models during server-side training, which introduces extra risk as the communication can be intercepted and the server can be attacked.

% On the contrary, FedSDD does not require access to the raw client models since it utilizes multiple aggregated models to build the ensemble, which is compatible to the techniques like secure aggregation, where the individual client model updates or weights do not need to be  disclosed to the server. This reduces the chance of being attacked during communication or server storage, and achieves the same privacy level as weight averaging based methods. To summarise, FedSDD is able to achieve a higher level of privacy protection than existing distillation based methods.

\subsection{Parallel server-side and client-side training of FedSDD}
\label{subsec:parallel}

In practical federated learning systems, clients are not always available for training. When there are limited number of available clients, the parallelism introduced in FedSSD can significantly reduce the round time by performing the server-side KD and the client-side local training simultaneously. In particular, the training of the global model $k$, with $k > 0$, i.e. except the main global model, in the $t$-th round can start, if its $t-1$-th round training is completed. This independence can also be observed from Figure 1 in the main paper, which shows that the operation in each group only depends on the previous process within the same group.

\begin{figure}
	\centering
	\begin{subfigure}[b]{1.00\textwidth}
		\centering
		\includegraphics[width=\textwidth]{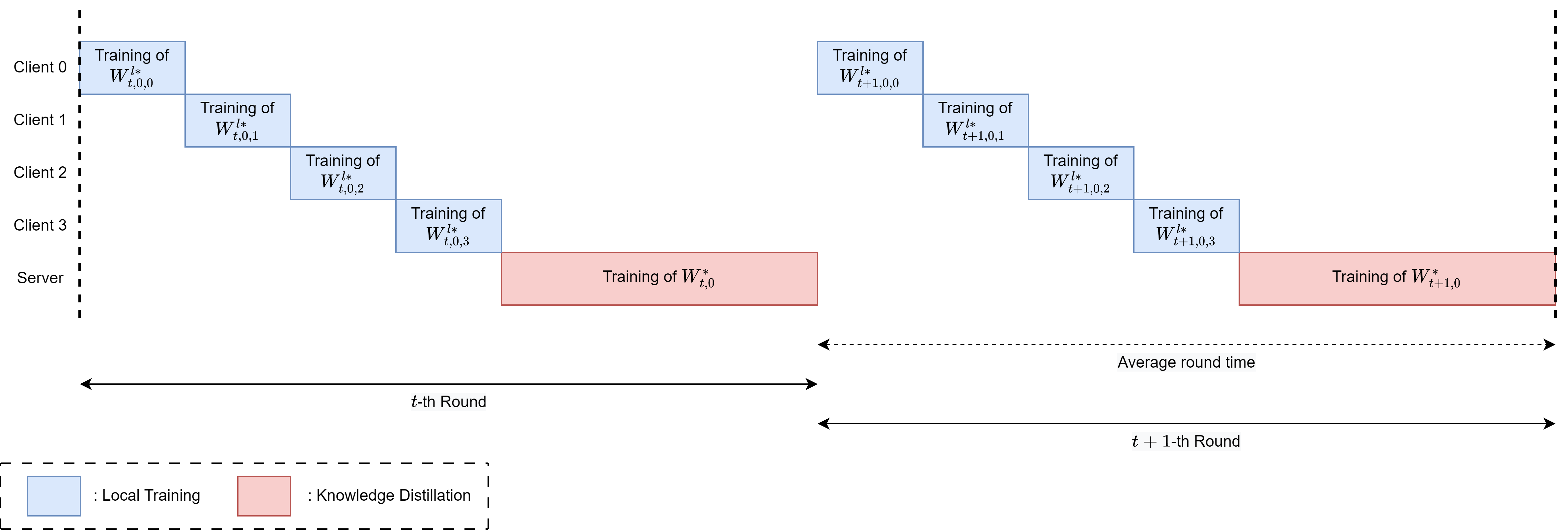}
		\caption{Existing Methods (e.g. FedDF)}
	\end{subfigure}
	\begin{subfigure}[b]{0.7923\textwidth}
		\centering
		\includegraphics[width=\textwidth]{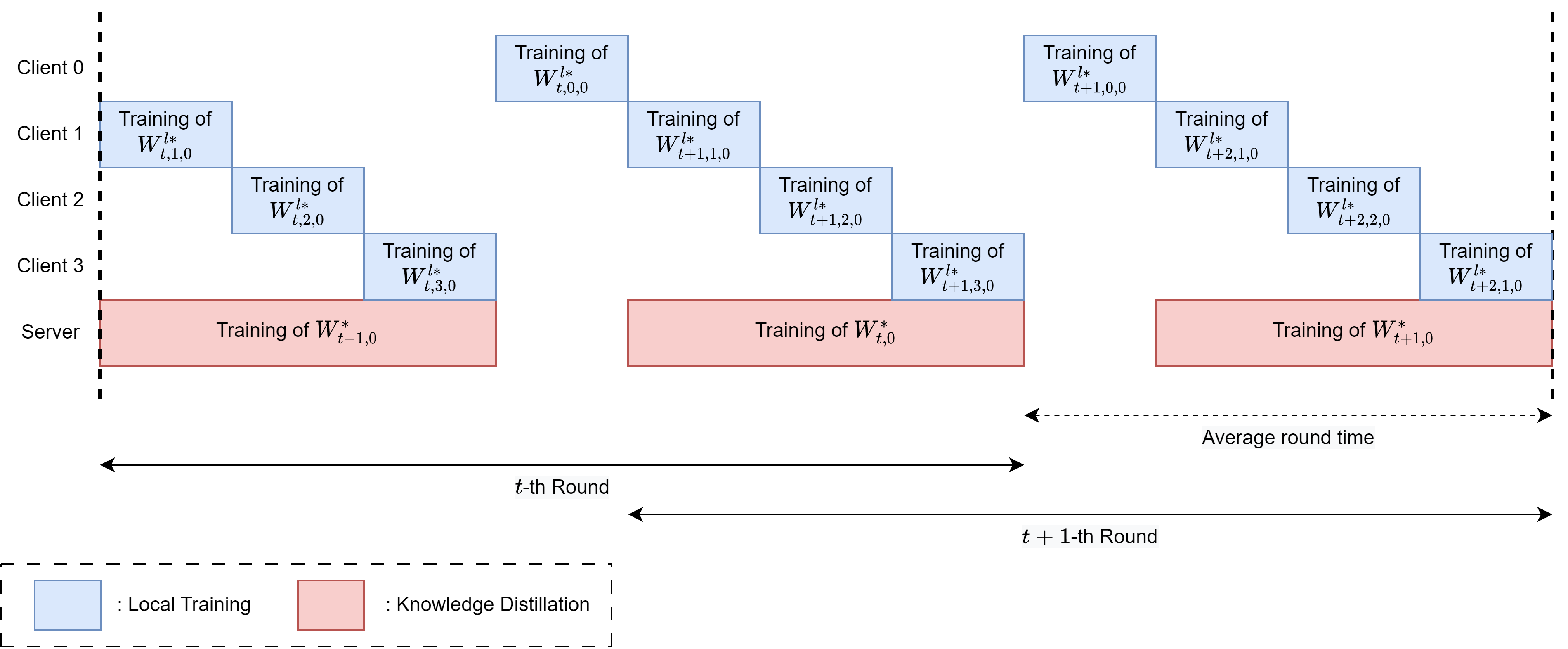}
		\caption{FedSDD}
	\end{subfigure}
	\caption{An example for the training scheduling comparison between FedSDD and other methods. When clients are not always available, the parallelism introduced in FedSSD can reduce the round time.}
	\label{fig:parallelism_uniform}
\end{figure}

In Figure \ref{fig:parallelism_uniform} of this supplementary material, we demonstrate one scheduling example for FedSDD and compare it with other distillation based methods, e.g. FedDF \cite{lin2020ensemble}, to illustrate the benefits of parallelism.  In this example, we assume there are four clients, and only one is available at any given time. For the sake of simplicity, we further assume there are four global models for FedSSD and the main global model is always assigned to the first client. Under such circumstances, existing methods need to perform KD and local training sequentially, while FedSSD can perform them simultaneously to reduce the round time.

%\newpage

\end{document}